# Generalization of Clauses under Implication

**Peter Idestam-Almquist**　　　　　　　　　　　　　　　　　　　　　　PI@DSV.SU.SE
*Department of Computer and Systems Sciences, Stockholm University*
*Electrum 230, S-164 40 Kista, Sweden*

## Abstract

In the area of inductive learning, generalization is a main operation, and the usual definition of induction is based on logical implication. Recently there has been a rising interest in clausal representation of knowledge in machine learning. Almost all inductive learning systems that perform generalization of clauses use the relation $\theta$-subsumption instead of implication. The main reason is that there is a well-known and simple technique to compute least general generalizations under $\theta$-subsumption, but not under implication. However generalization under $\theta$-subsumption is inappropriate for learning recursive clauses, which is a crucial problem since recursion is the basic program structure of logic programs.

We note that implication between clauses is undecidable, and we therefore introduce a stronger form of implication, called T-implication, which is decidable between clauses. We show that for every finite set of clauses there exists a least general generalization under T-implication. We describe a technique to reduce generalizations under implication of a clause to generalizations under $\theta$-subsumption of what we call an expansion of the original clause. Moreover we show that for every non-tautological clause there exists a T-complete expansion, which means that every generalization under T-implication of the clause is reduced to a generalization under $\theta$-subsumption of the expansion.

## 1. Introduction

The topic of this paper is generalization of clauses, which is a central problem in the area of Inductive Logic Programming (ILP) (Muggleton, 1991, 1993). ILP can be seen as the intersection of inductive machine learning and computational logic. In inductive machine learning the goal is to develop techniques for inducing hypotheses from examples (observations). By using the rich representation formalism of computational logic (clauses) for hypotheses and examples, ILP can overcome the limitations of classical machine learning representations, such as decision trees (Quinlan, 1986).

By using a clausal representation we have the ability to learn all types of hypotheses describable in first-order logic, in particular the important class of recursive hypotheses. Another advantage of using a clausal representation is that clausal theories are easy to manipulate for machine learning algorithms. This is due to that changes to a clausal theory by adding or deleting clauses or literals have clear and simple effects on the generality of the theory. The reader is referred to two introductions to ILP, one presented by Muggleton and De Raedt (1994), and one by Lavrač and Džeroski (1994). Lavrač and De Raedt (1995) present a recent survey of ILP research.

We use the following definition of induction. A theory (background knowledge) $T$, a set of positive examples $\{E_1^+, \ldots, E_n^+\}$ and a set of negative examples $\{E_1^-, \ldots, E_m^-\}$ of a target concept are given. Then a hypothesis $H$ for the target concept is an *inductive conclusion* if and only if:

1) $T \not\models E_1^+ \wedge \ldots \wedge E_n^+$,

2) $T \wedge H \models E_1^+ \wedge \ldots \wedge E_n^+$, and

3) $T \wedge H \not\models E_1^- \vee \ldots \vee E_m^-$.

In other words, the positive examples should not be a logical consequence of the theory alone, but a logical consequence of the theory together with the hypothesis, and no negative example should be a logical consequence of the theory and the hypothesis. Using clausal representation $T$, $H$, $\{E_1^+, \ldots, E_n^+\}$ and $\{E_1^-, \ldots, E_m^-\}$ are sets of clauses.

In this paper we concentrate on the subproblem in inductive learning of finding a clause that is a generalization of a set of positive examples. In other words, finding a clause $C$ such that
$$C \models E_1^+ \wedge \ldots \wedge E_n^+.$$
We are particularly interested in least general generalizations, since every generalization of a set of clauses is also a generalization of the least general generalization of this set of clauses. Therefore a least general generalization in some sense represents all generalizations. A least general generalization is also consistent with the negative examples whenever there exists a consistent generalization.

The most natural and straightforward basis for generalization is implication, since induction is defined in terms of logical consequence. Plotkin has described (1970, 1971a) a technique for the computation of least general generalizations of clauses under a relation called $\theta$-subsumption. This relation has been accorded much interest, and it is often used instead of implication, since it is easier to compute. However, there is a difference between $\theta$-subsumption and implication, which sometimes causes the generalizations obtained by Plotkin's technique to be over-generalizations with respect to implication.

Consider the following clauses in which $s$ denotes the successor function:
$$\begin{aligned} C_1 &= (\ number(s(0)) \leftarrow number(0)\ ), \\ C_2 &= (\ number(s^3(0)) \leftarrow number(s(0))\ ), \\ D_1 &= (\ number(s(x)) \leftarrow number(y)\ ), \text{ and} \\ D_2 &= (\ number(s(x)) \leftarrow number(x)\ ). \end{aligned}$$

The clause $D_1$ is a least general generalization under $\theta$-subsumption (LGG$\theta$) of $C_1$ and $C_2$, and the clause $D_2$ is a least general generalization under implication (LGGI) of $C_1$ and $C_2$. It is clear that $D_1$ is strictly more general than $D_2$, both under $\theta$-subsumption and under implication. It is also clear that $D_2$ is more appropriate in a definition of natural number.

To learn recursive clauses, generalization under $\theta$-subsumption is not very adequate, as illustrated above. The ability to learn recursive clauses is crucial, since recursion is the basic program structure of logic programs.

In section 2, we describe the most important results concerning generalization under $\theta$-subsumption, and present a theoretically study of generalization under implication. In section 3, we present a technique to reduce implication to $\theta$-subsumption based on or-introduction of literals. Finally, our results, computational complexity and future work are discussed in section 4.

We assume the reader to be familiar with the basic notions and notations in Logic Programming (Lloyd, 1987) and/or Automatic Theorem Proving (Chang & Lee, 1973; Gallier, 1986).

## 2. Generalization of Clauses

In the area of Inductive Logic Programming (ILP), the framework for generalization of clauses developed by Plotkin (1970, 1971b, 1971a), has been accorded much interest. In this section we will describe this framework, which is based on a relation known as $\theta$-subsumption, and the most important results connected with it.

Since generalization under $\theta$-subsumption is not sufficient for generalization of recursive clauses, as shown in the introduction, we will study the theory of generalization under implication. We note that implicaton between clauses is undecidable, and we will therefore introduce a restricted form of implication, called T-implication.

### 2.1 Generalization under $\theta$-subsumption

**Definition** A clause $C$ $\theta$-*subsumes* a clause $D$, denoted $C \preceq D$, if and only if there exists a substitution $\theta$ such that $C\theta \subseteq D$. Two clauses $C$ and $D$ are *equivalent under $\theta$-subsumption*, denoted $C \sim D$, if and only if $C \preceq D$ and $D \preceq C$.

$\theta$-subsumption is reflexive and transitive. Two clauses may be equivalent under $\theta$-subsumption without being variants. Two clauses $C$ and $D$ are *variants*, denoted $C \simeq D$, if they are equal up to variable renaming.

**Example** Consider the following clauses:
$$C = (\ p(x) \leftarrow q(x,y),\ q(y,z),\ q(z,w),\ q(w,x)\ ),$$
$$D = (\ p(x) \leftarrow q(x,y),\ q(y,x),\ q(x,x)\ ),\ \text{and}$$
$$E = (\ p(x) \leftarrow q(x,x)\ ).$$

We have $C \preceq D$ since $C\{z/x, w/y\} \subseteq D$, $D \preceq E$ since $D\{y/x\} \subseteq E$, and thus $C \preceq E$. We also have $E \preceq D$. Hence, $D \sim E$ and still $D \not\simeq E$.

Theorem 1 states that $\theta$-subsumption between clauses is decidable. This was first shown by Robinson (1965, page 39).

**Theorem 1 (Decidability of $\theta$-subsumption between clauses)** *Let $C$ and $D$ be clauses. Then there exists a procedure to decide if $C \preceq D$.*

As mentioned in the introduction, we are particularly interested in least general generalizations. The main reason is that a least general generalization includes the information of all consistent generalizations.

**Definition** A clause $C$ is a *generalization under $\theta$-subsumption* of a set of clauses $S = \{D_1, \ldots, D_n\}$ if and only if, for every $1 \leq i \leq n$, $C \preceq D_i$. A generalization under $\theta$-subsumption $C$ of $S$ is a *least general generalization under $\theta$-subsumption* (LGG$\theta$) of $S$ if and only if, for every generalization under $\theta$-subsumption $C'$ of $S$, $C' \preceq C$.

**Example** Consider the following clauses:
$$C = (\ p(a) \leftarrow q(a),\ q(b)\ ),$$
$$D = (\ p(b) \leftarrow q(b),\ q(x)\ ),$$
$$E = (\ p(y) \leftarrow q(y),\ q(b)\ ),\ \text{and}$$
$$F = (\ p(y) \leftarrow q(y),\ q(b),\ q(z),\ q(w)\ ).$$

Both clauses $E$ and $F$ are LGG$\theta$s of $\{C, D\}$.

In general, an LGG$\theta$ is not unique, as shown by the example above. However, it is unique up to $\theta$-subsumption equivalence. Plotkin has shown (1971a, page 82) that there exists an LGG$\theta$ of every finite set of clauses.

**Theorem 2 (Existence of LGG$\theta$s)** *Let S be a finite set of clauses. Then there exists an LGG$\theta$ of S.*

An LGG$\theta$ of a set of clauses is computable, and Plotkin has described (1971a) an algorithm for that. This algorithm is quite simple and easy to implement, but computationally expensive.

## 2.2 Generalization under Implication

Implication is the most natural and straightforward basis for generalization in inductive learning, since the concept of induction can be defined as the inverse of logically entailment.

**Definition** A clause $C$ *implies* a clause $D$, denoted $C \Rightarrow D$, if and only if every model for $C$ is a model for $D$ ($\{C\} \models D$). Two clauses $C$ and $D$ are *equivalent under implication*, denoted $C \Leftrightarrow D$, if and only if $C \Rightarrow D$ and $D \Rightarrow C$.

It is well-known that implication is reflexive and transitive. Two clauses may be equivalent under implication without being equivalent under $\theta$-subsumption.

**Example** Consider the following clauses:
$$C = (\ p(x,y,z) \leftarrow p(y,z,x)\ ), \text{ and}$$
$$D = (\ p(x,y,z) \leftarrow p(z,x,y)\ ).$$

Then we have $C \Leftrightarrow D$, since $D$ is a resolvent of $C$ resolved with itself, and $C$ is a resolvent of $D$ resolved with itself. We also have $C \not\supseteq D$, and even $C \not\succeq D$.

It has been claimed that implication and $\theta$-subsumption are equivalent for function-free clauses (Helft, 1987). This is wrong as shown by the example above. The above example also shows that if a clause $C$ implies a clause $D$ then $C$ does not necessarily $\theta$-subsume $D$. It is well-known that implication is a strictly weaker relation between clauses than $\theta$-subsumption.

**Proposition 3** *Let $C$ and $D$ be two clauses. If $C \preceq D$ then $C \Rightarrow D$.*

Proposition 3 has been proved by Idestam-Almquist (1993a, page 21). Unfortunately implication between clauses is problematic since it is undecidable, which has been proved by Schmidt-Schauss (1988, page 294).

**Theorem 4 (Undecidability of implication between clauses)** *Let $C$ and $D$ be clauses. Then there exists no procedure to decide if $C \Rightarrow D$.*

Niblett (1988) has claimed that implication between Horn clauses is decidable. This result has later been proved to be false (Marcinkowski & Pacholski, 1992).

The definition of a least general generalization under implication (LGGI) follows the definition of an LGG$\theta$.

**Definition** A clause $C$ is a *generalization under implication* of a set of clauses $S = \{D_1, \ldots, D_n\}$ if and only if, for every $1 \leq i \leq n$, $C \Rightarrow D_i$. A generalization under implication $C$ of $S$ is a *least general generalization under implication* (LGGI) of $S$ if and only if, for every generalization under implication $C'$ of $S$, $C' \Rightarrow C$.

**Example** Consider the following clauses:

$$C = (\ p(f(a)) \leftarrow p(a)\ ),$$
$$D = (\ p(f^2(b)) \leftarrow p(b)\ ),$$
$$E = (\ p(f(x)) \leftarrow p(y)\ ), \text{ and}$$
$$F = (\ p(f(z)) \leftarrow p(z)\ ).$$

The clause $E$ is an LGG$\theta$ of $\{C, D\}$, and $F$ is an LGGI of $\{C, D\}$. The LGG$\theta$ (clause $E$) is strictly more general than the LGGI (clause $F$), both under implication and under $\theta$-subsumption, since $E \Rightarrow F$ but $F \not\Rightarrow E$, and $E \preceq F$ but $F \not\preceq E$.

Whether there exists an LGGI of every finite set of clauses is still an open problem. However, since implication between clauses is undecidable, it is clear that in general an LGGI is not computable.

### 2.3 T-implication

Because implication between clauses is undecidable, we here introduce a stronger form of implication called T-implication, which is decidable between clauses. It is called T-implication since it is defined w.r.t. a finite set of ground terms $T$. In our presentation we use the notions of instance set of clauses, Skolem substitution, and term set of sets of clauses.

**Definition** Let $C$ be a clause, $\{x_1, \ldots, x_n\}$ the set of variables in $C$, and $T$ a set of terms. Then the *instance set* $\mathcal{I}(C, T)$ of $C$ w.r.t. $T$ is $\{C\theta \mid \theta = \{x_1/t_1, \ldots, x_n/t_n\}$ where $\{t_1, \ldots, t_n\} \subseteq T\}$.

**Definition** Let $\sigma$ be a substitution, $C$ a clause, $\{x_1, \ldots, x_n\}$ the set of variables occurring in $C$, $S$ a set of clauses, and $F$ the set of function symbols occurring in $S \cup \{C\}$. Then $\sigma$ is a *Skolem substitution* for $C$ w.r.t. $S$ if and only if $\{x_1/a_1, \ldots, x_n/a_n\} \subseteq \sigma$ where $a_1, \ldots, a_n$ are distinct constants, and $F \cap \{a_1, \ldots, a_n\} = \emptyset$.

**Definition** Let $\{D_1, \ldots, D_n\}$ be a set of clauses such that $D_1, \ldots, D_n$ have no variables in common, $S$ be a set of clauses, $\sigma$ a substitution, and $T$ a set of terms. Then $T$ is a *term set* of $\{D_1, \ldots, D_n\}$ by $\sigma$ w.r.t. $S$ if and only if:
a) $\sigma$ is a Skolem substitution for $\{D_1, \ldots, D_n\}$ w.r.t. $S$, and
b) $T$ is finite and includes all terms and subterms occurring in $\{D_1, \ldots, D_n\}\sigma$.
If $T$ is equal to the set of terms and subterms occurring in $\{D_1, \ldots, D_n\}\sigma$ then $T$ is a *minimal term set* of $\{D_1, \ldots, D_n\}$ by $\sigma$ w.r.t. $S$.

**Definition** Let $C$ and $D$ be clauses, and $T$ a term set of $\{D\}$ by $\sigma$ w.r.t. $\{C\}$. Then $C$ *T-implies* $D$ w.r.t. $T$, denoted $C \Rightarrow_T D$, if and only if $\mathcal{I}(C, T) \models D\sigma$. Two clauses $C$ and $D$ are *equivalent under T-implication* w.r.t. $T'$, denoted $C \Leftrightarrow_{T'} D$, if and only if $C \Rightarrow_{T'} D$ and $D \Rightarrow_{T'} C$, where $T'$ is a term set of $\{C, D\}$.

Note that the definition of T-implication is independent of the choice of the Skolem substitution $\sigma$. In the following, if we say that a clause $C$ T-implies a clause $D$ without explicitly stating $T$, we mean that $C$ T-implies $D$ w.r.t. a minimal term set of $\{D\}$. Note that if $C$ T-implies $D$ w.r.t. a minimal term set of $D$ then $C$ T-implies $D$ w.r.t. any term set of $D$.

**Example** Consider the following clauses $C$ and $D$, substitution $\sigma$, set of terms $T$ and set of clauses $\mathcal{I}(C, T)$:
$$C = (\, p(f(x)) \leftarrow p(x)\, ),$$
$$D = (\, p(f^2(y)) \leftarrow p(y)\, ),$$
$$\sigma = \{y/a\},$$
$$T = \{a, f(a), f^2(a)\}, \text{ and}$$
$$\mathcal{I}(C, T) = \{\, (\, p(f(a)) \leftarrow p(a)\, ),$$
$$(\, p(f^2(a)) \leftarrow p(f(a))\, ),$$
$$(\, p(f^3(a)) \leftarrow p(f^2(a))\, )\,\}.$$

Then $T$ is a minimal term set of $\{D\}$ by $\sigma$ w.r.t. $\{C\}$, and $\mathcal{I}(C, T)$ is the instance set of $C$ w.r.t. $T$. We have that $\mathcal{I}(C, T) \models D\sigma$ and thus $C \Rightarrow_T D$. Note that $C \Rightarrow D$, and that $C \not\preceq D$.

Like implication, T-implication is reflexive, but unlike implication, T-implication is not transitive (Idestam-Almquist, 1993a). The relationship between implication and T-implication, described in Corollary 6 below, follows from Herbrand's theorem. For a proof of Herbrand's theorem the reader is referred to a book by Chang and Lee (1973, page 61). In our proof of Corollary 6 we use the notion of the complement of a clause.

**Definition** Let $C = (A_1, \ldots, A_m \leftarrow B_1, \ldots, B_n)$ be a clause, $T$ a set of clauses, and $\sigma = \{x_1/a_1, \ldots, x_k/a_k\}$ a Skolem substitution for $C$ w.r.t. $T$. Then the set of ground unit clauses $\{(\leftarrow A_1)\sigma, \ldots, (\leftarrow A_m)\sigma, (B_1 \leftarrow)\sigma, \ldots, (B_n \leftarrow)\sigma\}$ is the *complement* $\overline{C}$ of $C$ by $\sigma$ w.r.t. $T$.

**Theorem 5 (Herbrand's theorem)** *A set of clauses $S$ is unsatisfiable if and only if there exists a finite unsatisfiable set $S'$ of ground instances of clauses in $S$.*

**Corollary 6 (Relationship between implication and T-implication)** *Let $C$ and $D$ be clauses. Then:*
*a) if $C \Rightarrow_T D$ for some term set $T$ of $\{D\}$ then $C \Rightarrow D$, and*
*b) if $C \Rightarrow D$ then there exists a term set $T$ of $\{D\}$ such that $C \Rightarrow_T D$.*

**Proof:** a) If $C \Rightarrow_T D$ then $\mathcal{I}(C, T) \models D\sigma$, where $T$ is a term set of $\{D\}$ by $\sigma$ w.r.t. $\{C\}$. Hence, $\mathcal{I}(C, T) \cup \overline{D} \models \bot$, where $\overline{D}$ is the complement of $D$ by $\sigma$. By Theorem 5, $\{C\} \cup \overline{D} \models \bot$, and thus $C \Rightarrow D$.

b) If $C \Rightarrow D$ then $\{C\} \cup \overline{D} \models \bot$, where $\overline{D}$ is the complement of $D$ by $\sigma$ w.r.t. $\{C\}$. Then by Theorem 5, there exists a term set $T$ of $\{D\}$ such that $\mathcal{I}(C, T) \cup \overline{D} \models \bot$, and thus $\mathcal{I}(C, T) \models D\sigma$. Then by definition $C \Rightarrow_T D$. □

It follows from Corollary 6 that T-implication can become an arbitrary good approximation of implication by extending the considered term set. T-implication is a strictly stronger relation between clauses than implication. The following example illustrates that if a clause $C$ implies a clause $D$ then $C$ does not necessarily T-imply $D$.

**Example** Consider the following clauses $C$, $D$ and $E$, and set of terms $T$:

$$C = (\ p(f(x), y) \leftarrow p(z, x)\ ),$$
$$D = (\ p(f(x), y) \leftarrow p(z, w)\ ),$$
$$E = (\ p(f(a), a) \leftarrow p(a, f(a))\ ),\text{ and}$$
$$T = \{a, f(a)\}.$$

Then $C \Rightarrow E$ since $D$ is a resolvent of $C$ resolved with itself and $E$ is an instance of $D$. The set of terms $T$ is a minimal term set of $E$. We do not show here the whole set $\mathcal{I}(C, T)$, but just point out that $\mathcal{I}(C, T) \not\models E$ and thus $C \not\Rightarrow_T E$. However if we extend $T$ to $T' = \{a, f(a), f^2(a)\}$ then $\mathcal{I}(C, T') \models E$, and thus $C \Rightarrow_{T'} E$.

Below we show that if a clause $C$ $\theta$-subsumes a clause $D$ then $C$ also T-implies $D$. Thus, T-implication is a strictly weaker relation between clauses than $\theta$-subsumption. We also show decidability of T-implication between clauses.

**Proposition 7** *Let $C$ and $D$ be clauses and $T$ a term set of $\{D\}$. If $C \preceq D$ then $C \Rightarrow_T D$.*

**Proof:** If $C \preceq D$ then there exists a substitution $\theta$ such that $C\theta \subseteq D$. Let $T$ be a term set of $\{D\}$ by $\sigma$ w.r.t. $\{C\}$. Then we have $C\theta\sigma \in \mathcal{I}(C, T)$. We also have $C\theta\sigma \subseteq D\sigma$, and thus $C\theta\sigma \preceq D\sigma$. Then by Proposition 3, $C\theta\sigma \Rightarrow D\sigma$ ($\{C\theta\sigma\} \models D\sigma$). Consequently $\mathcal{I}(C, T) \models D\sigma$, and then by definition $C \Rightarrow_T D$. □

**Theorem 8 (Decidability of T-implication between clauses)** *Let $C$ and $D$ be clauses and $T$ a term set of $\{D\}$. Then there exists a procedure to decide if $C \Rightarrow_T D$.*

**Proof:** By the definition of T-implication we have $\mathcal{I}(C, T) \models D\sigma$ where $T$ is a term set of $D$ by $\sigma$ w.r.t. $\{C\}$. We have that $\mathcal{I}(C, T)$ is a set of ground clauses and $D\sigma$ is a ground clause. Thus, it follows from the decidability of logical consequence in propositional logic that T-implication is decidable. □

### 2.4 Generalization under T-implication

A least general generalization under T-implication (LGGT) is defined similar to an LGG$\theta$ and an LGGI.

**Definition** Let $C$ be a clause, $S = \{D_1, \ldots, D_n\}$ a set of clauses, and $T$ a term set of $S$ w.r.t. $\{C\}$. Then $C$ is a *generalization under T-implication* of $S$ w.r.t. $T$ if and only if, for every $1 \leq i \leq n$, $C \Rightarrow_T D_i$. A generalization under T-implication $C$ of $S$ w.r.t. $T$ is a *least general generalization under T-implication* (LGGT) of $S$ w.r.t. $T$ if and only if, for every generalization under T-implication $C'$ of $S$ w.r.t. $T$, $C' \Rightarrow_{T'} C$, where $T'$ is a minimal term set of $C$.

**Example** Consider the following clauses:

$$C = (\ p(f(a)) \leftarrow p(a)\ ),$$
$$D = (\ p(f^2(b)) \leftarrow p(b)\ ),$$
$$E = (\ p(f(x)) \leftarrow p(y)\ ),\ \text{and}$$
$$F = (\ p(f(z)) \leftarrow p(z)\ ).$$

The clause $E$ is an LGG$\theta$ of $\{C, D\}$, and $F$ is both an LGGI and an LGGT of $\{C, D\}$. The LGGT is strictly more specific than the LGG$\theta$, since $E \Rightarrow F$ and $F \not\Rightarrow E$.

Below we prove that there exists an LGGT of every finite set of clauses. In fact we prove something stronger, namely that there exists, what we call, a complete LGGT of every finite set of non-tautological clauses. Note that a complete LGGT is $\theta$-subsumed by any other generalization under T-implication.

**Definition** Let $C$ be an LGGT of a set of clauses $S$ w.r.t. a term set $T$. Then $C$ is a *complete LGGT* of $S$ w.r.t. $T$ if and only if, for every generalization under T-implication $C'$ of $S$ w.r.t. $T$, $C' \preceq C$.

If $C$ is a clause then we let $C^+$ denote the set of positive literals in $C$, and $C^-$ the set of negative literals in $C$. The following proposition has been proved by Gottlob (1987, page 110).

**Proposition 9** *Let $C = C^+ \cup C^-$ be a clause and $D = D^+ \cup D^-$ a non-tautological clause. If $C \Rightarrow D$ then $C^+ \preceq D^+$ and $C^- \preceq D^-$.*

**Lemma 10** *Let $S$ be a finite set of non-tautological clauses, $T = \{t_1, \ldots, t_m\}$ a term set of $S$, $V = \{x_1, \ldots, x_m\}$ a set of variables, and $G = \{C_1, C_2, \ldots\}$ the (possibly infinite) set of all generalizations under T-implication of $S$ w.r.t. $T$. Then the set $G' = \mathcal{I}(C_1, V) \cup \mathcal{I}(C_2, V) \cup \ldots$ is a finite set of clauses.*

**Proof:** Let $d$ be the maximal depth of a clause in $S$, and $F_S$ and $F_G$ the sets of predicate and function symbols occurring in the clauses in $S$ and $G$ respectively. Then $F_G \cup V$ is the set of variables, predicate and function symbols occurring in the clauses in $G'$. By Corollary 6, $G$ is a set of generalizations under implication of $S$. Then, by Proposition 9 and the definition of $\theta$-subsumption, $F_G \subseteq F_S$ and the maximal depth of a clause in $G$ is $d$. Hence $F_G \cup V$ is finite and the maximal depth of a clause in $G'$ is $d$, and consequently $G'$ is a finite set of clauses. □

**Lemma 11** *Let $C$ be a clause, $S$ a set of clauses, $V = \{x_1, \ldots, x_m\}$ a set of variables, and $T = \{t_1, \ldots, t_m\}$ a term set of $S$ by $\sigma$ w.r.t $\{C\}$, such that $C$ is a generalization under T-implication of $S$ w.r.t. $T$. Then there exists an LGG$\theta$ $E$ of $\mathcal{I}(C, V)$ such that $E$ is a generalization under T-implication of $S$ w.r.t. $T$.*

**Proof:** Let $\mathcal{I}(C, V) = \{C\rho_1, \ldots, C\rho_k\}$. Then, $\rho_1, \ldots, \rho_k$ are variable-pure substitutions, and for every LGG$\theta$ $F$ of $\mathcal{I}(C, T)$ and every $1 \leq i \leq k$, we have $C \preceq F$ and $F \preceq C\rho_i$. Then, there exists an LGG$\theta$ $E$ of $\mathcal{I}(C, T)$ and variable-pure substitutions $\theta_1, \ldots, \theta_k$ such that, for every $1 \leq i \leq k$, $E\theta_i \subseteq C\rho_i$.

Let $\gamma = \{x_1/t_1, \ldots, x_m/t_m\}$, and then $\mathcal{I}(C,T) = \{C\rho_1\gamma, \ldots, C\rho_k\gamma\}$. Since $C$ is a generalization under T-implication of $S$ w.r.t. $T$, we have $\{C\rho_1\gamma, \ldots, C\rho_k\gamma\} \models S\sigma$. For every $1 \leq i \leq k$, $E\theta_i\gamma \preceq C\rho_i\gamma$. Then, for every $1 \leq i \leq k$, by Proposition 3, $E\theta_i\gamma \Rightarrow C\rho_i\gamma$, and thus $\{E\theta_1\gamma, \ldots, E\theta_k\gamma\} \models \{C\rho_1\gamma, \ldots, C\rho_k\gamma\}$. Since $\theta_1, \ldots, \theta_k$ are variable-pure substitutions, we have $\{E\theta_1\gamma, \ldots, E\theta_k\gamma\} \subseteq \mathcal{I}(E,T)$. Thus, $\mathcal{I}(E,T) \models \{C\rho_1\gamma, \ldots, C\rho_k\gamma\}$, and $\mathcal{I}(E,T) \models S\sigma$. Consequently $E$ is a generalization under T-implication of $S$ w.r.t. $T$. □

**Lemma 12** *Let $C$, $D$ and $E$ be clauses such that $C$ and $D$ have no variables in common, and let $T$ be a term set of $\{E\}$ by $\sigma$ w.r.t. $\{C, D\}$. If $C \Rightarrow_T E$ and $D \Rightarrow_T E$ then $C \cup D \Rightarrow_T E$.*

**Proof:** If $C \Rightarrow_T E$ and $D \Rightarrow_T E$ then by definition $\mathcal{I}(C,T) \models E\sigma$ and $\mathcal{I}(D,T) \models E\sigma$. Let $\mathcal{I}(C,T) = \{C_1, \ldots, C_n\}$ and $\mathcal{I}(D,T) = \{D_1, \ldots, D_m\}$. Then $\mathcal{I}(C \cup D, T) = \{C_i \cup D_j \mid 1 \leq i \leq n \text{ and } 1 \leq j \leq m\}$. Let $I$ be a model for $\mathcal{I}(C \cup D, T)$. Then, for every $1 \leq i \leq n$ and $1 \leq j \leq m$, $I$ is a model for $C_i \cup D_j$. Hence if $I$ is not a model for $C_i$ for some $1 \leq i \leq n$ then $I$ must be a model for $D_j$ for every $1 \leq j \leq m$. Then it follows that either $I$ is a model for $\mathcal{I}(C,T)$ or $I$ is a model for $\mathcal{I}(D,T)$, and thus $I$ is a model for $E\sigma$. Consequently, $\mathcal{I}(C \cup D, T) \models E\sigma$, and then by definition $C \cup D \Rightarrow_T E$. □

**Theorem 13 (Existence of complete LGGTs)** *Let $S$ be a finite set of non-tautological clauses, and $T$ a term set of $S$. Then there exists a complete LGGT of $S$ w.r.t. $T$.*

**Proof:** Let $T = \{t_1, \ldots, t_m\}$, $V = \{x_1, \ldots, x_m\}$ be a set of variables, and $G = \{C_1, C_2, \ldots\}$ the (possibly infinite) set of all generalizations under T-implication of $S$ w.r.t. $T$. By Lemma 10, the set $G' = \mathcal{I}(C_1, V) \cup \mathcal{I}(C_2, V) \cup \ldots$ is a finite set of clauses. Since $G'$ is finite, the set $\{\mathcal{I}(C_1, V), \mathcal{I}(C_2, V), \ldots\}$ is also finite.

For every $i \geq 1$, by Lemma 11, there exists an LGG$\theta$ $E_i$ of $\mathcal{I}(C_i, V)$ such that $E_i$ is a generalization under T-implication of $S$ w.r.t. $T$. Then rename the variables in $E_1, E_2, \ldots$ such that, for every $k \geq 1$ and $p \geq 1$, $E_k = E_p$ whenever $\mathcal{I}(C_k, V) = \mathcal{I}(C_p, V)$ and otherwise $E_k$ and $E_p$ have no variables in common. Then the set $\{E_1, E_2, \ldots\}$ is finite, since $\{\mathcal{I}(C_1, V), \mathcal{I}(C_2, V), \ldots\}$ is finite. Let $F = E_1 \cup E_2 \cup \ldots$, which consequently is a clause.

For every $i \geq 1$, by the definition of an LGG$\theta$, $C_i \preceq E_i$, and thus $C_i \preceq F$. Then, for every $i \geq 1$, by Proposition 7, $C_i \Rightarrow_T F$. As showed above, for every $i \geq 1$, $E_i$ is a generalization under T-implication of $S$ w.r.t. $T$. Then, by Lemma 12, $F$ is a generalization under T-implication of $S$ w.r.t. $T$. Consequently, $F$ is a complete LGGT of $S$ w.r.t. $T$. □

**Theorem 14 (Existence of LGGTs)** *Let $S$ be a finite set of clauses, and $T$ a term set of $S$. Then there exists an LGGT of $S$ w.r.t. $T$.*

**Proof:** Let $D$ be a tautology and $\sigma$ a Skolem substitution for $D$. Then $\top \models D\sigma$, and thus for every clause $C$, $C \Rightarrow_T D$. If every clause in $S$ is a tautology, then every clause is a generalization under T-implication of $S$ w.r.t. $T$, and every tautology is an LGGT of $S$ w.r.t. $T$. Let $S'$ be the set of clauses obtained from $S$ by removing all tautologies. It is clear that every generalization under T-implication of $S$ w.r.t. $T$ also is a generalization under T-implication of $S'$ w.r.t. $T$. By Theorem 13, there exists an LGGT of $S'$ w.r.t. $T$. Consequently there exists an LGGT of $S$. □

IDESTAM-ALMQUIST## 3. Reduction of Implication to $\theta$-subsumption

There are generalizations under implication that are not generalizations under $\theta$-subsumption. Our main idea to find all generalizations under implication, is to reduce implication to $\theta$-subsumption, which can be achieved by inverting self-resolution. In this section we will describe a technique for inverting resolution based on or-introduction of literals. We will also introduce the notion of expansion of clauses, which summarizes our idea of reduction of implication to $\theta$-subsumption.

### 3.1 Difference between $\theta$-subsumption and Implication

In section 2.2, we showed that $C \Rightarrow D$ follows from $C \preceq D$, but not the converse. Hence, there are generalizations under implication that are not generalizations under $\theta$-subsumption. It follows from a result by Gottlob (1987) that the difference between $\theta$-subsumption and implication only concerns ambivalent clauses, as defined below.

**Definition** A clause $C$ is *ambivalent* if and only if there exist a positive literal $A \in C$ and a negative literal $B \in C$ such that $A$ and $B$ have the same predicate symbol.

**Example** The clause
$$C = (\ p(f^2(a)) \leftarrow q(b),\ p(a)\ )$$
is ambivalent since $p(f^2(a))$ and $\neg p(a)$ have the same predicate symbol. However, $C$ is not recursive since neither $p(a)$ nor $q(b)$ is unifiable with a variant of $p(f^2(a))$.

**Proposition 15** *Let $C$ be a clause and $D$ a non-ambivalent clause. Then $C \Rightarrow D$ if and only if $C \preceq D$.*

Proposition 15 has been proved by Gottlob (1987, page 110). It follows from this proposition that an LGG$\theta$ and an LGGI of a set of clauses, including at least one non-ambivalent clause, are equivalent.

Muggleton (1992) has investigated the relationship between resolution and implication between clauses. He describes the subsumption theorem (Lee, 1967) in terms of input resolution, and gives a corollary about the relationship between $\theta$-subsumption and implication between clauses. Unfortunately, this formulation of the subsumption theorem, which later also has been used by Idestam-Almquist (1993c, 1993a), has been shown to be wrong. Nienhuys-Cheng and de Wolf (1995) have given a counter-example which shows that the subsumption theorem for input resolution does not even hold in the special case where the considered set of clauses contains only one clause. Below we give the correct formulation of the subsumption theorem, which is based on the $n$th resolution (Robinson, 1965).

**Definition** A substitution $\theta$ is a *unifier* for a finite set of literals $S$ if and only if $S\theta$ is a singleton. A unifier $\theta$ for $S$ is a *most general unifier* (mgu) for $S$ if and only if for each unifier $\sigma$ of $S$ there exists a substitution $\gamma$ such that $\sigma = \theta\gamma$.

**Definition** Let $C$ be a clause, $\Gamma \subseteq C$ and $\gamma$ an mgu of $\Gamma$. Then $C\gamma$ is a *factor* of $C$.

476

**Definition** A clause $R$ is a *resolvent* of two clauses $C$ and $D$ if and only if there are $C\gamma$, $D\mu$, $A$, $B$ and $\theta$ such that:
a) $C\gamma$ is a factor of $C$ and $D\mu$ is a factor of $D$,
b) $C\gamma$ and $D\mu$ have no variables in common,
c) $A$ is a literal in $C\gamma$ and $B$ is a literal in $D\mu$,
d) $\theta$ is an mgu of $\{A, \overline{B}\}$, and
e) $R$ is the clause $((C\gamma - \{A\}) \cup (D\mu - \{B\}))\theta$.
The clauses $C$ and $D$ are called *parent clauses* of $R$.

**Definition** Let $T$ be a set of clauses. Then, the *$n$th resolution* of $T$, denoted $\mathcal{R}^n(T)$, is defined as:
a) $\mathcal{R}^0(T) = T$, and
b) $\mathcal{R}^n(T) = \mathcal{R}^{n-1}(T) \cup \{R \mid C, D \in \mathcal{R}^{n-1}(T)$ and $R$ is a resolvent of $C$ and $D\}$ if $n > 0$.

**Theorem 16 (Subsumption theorem)** *Let $T$ be a set of clauses and $C$ a non-tautological clause. Then $T \models C$ if and only if there exists a clause $D \in \mathcal{R}^n(T)$ such that $D \preceq C$ for some $n \geq 0$.*

Two different recent proofs of Theorem 16 have been presented, one by Nienhuys-Cheng and de Wolf (1995), and one by Bain and Muggleton (1992). There also exist at least two different earlier proofs of this theorem in the literature, one by Slagle, Chang and Lee (1969), and one by Kowalski (1970). We are interested in the number of resolutions involved in the computation of a clause, and therefore we introduce the notion of $n$th resolution layer. A clause in the $n$th resolution layer has been obtained from the original set of clauses by $n - 1$ resolutions.

**Definition** Let $T$ be a set of clauses. Then, the *$n$th resolution layer* of $T$, denoted $\mathcal{L}^n(T)$, is defined as:
a) $\mathcal{L}^1(T) = T$, and
b) $\mathcal{L}^n(T) = \{R \mid R$ is a resolvent of $C \in \mathcal{L}^m(T)$ and $D \in \mathcal{L}^p(T)$ where $m + p = n - 1$, $m \geq 1$ and $p \geq 1\}$ if $n > 1$.

**Corollary 17 (Implication between clauses using resolution)** *Let $C$ be a clause and $D$ a non-tautological clause. Then $C \Rightarrow D$ if and only if there exists a clause $E \in \mathcal{L}^n(\{C\})$ such that $E \preceq D$ for some $n \geq 1$.*

Corollary 17 follows from Theorem 16, and the observation that, for every $n \geq 1$, if a clause $C \in \mathcal{L}^n(T)$ then also $C \in \mathcal{R}^n(T)$. This corollary tells us that implication between clauses is equivalent to a combination of self-resolution and $\theta$-subsumption. Muggleton (1992) has introduced the notion of powers and roots of clauses for specializations and generalizations of clauses where the clauses are resolved with themselves. Below we present definitions of these and related concepts modified w.r.t. the correct definition of the subsumption theorem.

**Definition** A clause $D$ is an *$n$th power* of a clause $C$ if and only if $D$ is a variant of a clause in $\mathcal{L}^n(\{C\})$ ($n \geq 1$). We also say that $C$ is an *$n$th root* of $D$. A clause $D$ is an *indirect $n$th* *power* of a clause $C$ if and only if there exists a clause $E$ such that $E \preceq D$ and $E$ is an $n$th power of $C$. We also say that $C$ is an *indirect $n$th root* of $D$.

Let $C$ be a clause and $D$ an indirect $n$th power of $C$. Then $D$ is a *proper indirect $n$th power* of $C$ if and only if $C \not\preceq D$. We also say that $C$ is a *proper indirect $n$th root* of $D$.

**Example** Consider the following clauses:

$$C = (\ p(f(x)) \leftarrow p(x)\ ),$$
$$D = (\ p(f^2(x)) \leftarrow p(x)\ ),$$
$$E = (\ p(f^3(x)) \leftarrow p(x)\ ),$$
$$F = (\ p(f^2(a)) \leftarrow p(a),\ p(b)\ ),\ \text{and}$$
$$G = (\ p(x) \leftarrow p(b)\ ).$$

The clause $C$ is a second root of $D$, and a third root of $E$. The clause $C$ is also an indirect second root of $F$, since $C$ is a second root of $D$ and $D$ $\theta$-subsumes $F$. In fact $C$ is a proper indirect second root of $F$, since $C \not\preceq F$. For every $n \geq 1$, the clause $G$ is an indirect $n$th root of itself, but none of these indirect roots is a proper indirect root.

To say that a clause implies another non-tautological clause or to say that the clause is an indirect root of the other clause, is equivalent. However, to say that a clause is an indirect $n$th root for some specified $n$ is more informative.

Implication between clauses can be described as a combination of self-resolution and $\theta$-subsumption. Plotkin's algorithm to compute LGG$\theta$s gives us a suitable tool for finding generalizations under $\theta$-subsumption. Hence, to be able to find generalizations under implication we also need a technique to invert resolution.

### 3.2 Inverting One Resolution by Or-introduction

Other work on inverting resolution has primarily considered the problem of constructing one parent clause given the resolvent and the other parent clause (Muggleton & Buntine, 1988; Rouveirol & Puget, 1989; Wirth, 1989; Muggleton, 1990; Hume & Sammut, 1991; Idestam-Almquist, 1992; Rouveirol, 1992). Below we will describe how or-introduction can be used to construct two parent clauses from only the resolvent. Let $C$ and $D$ be clauses, and the following clause $R$ a resolvent of $C$ and $D$:

$$R = ((C\gamma - \{A\}) \cup (D\mu - \{B\}))\theta,$$

where $C\gamma$ is a factor of $C$, $D\mu$ is a factor of $D$, $A \in C\gamma$, $B \in D\mu$ and $\theta$ is an mgu for $\{A, \overline{B}\}$. We seek parent clauses of $R$ that are minimally general. Then we should let $\gamma$, $\mu$ and $\theta$ be empty substitutions, which corresponds to an assumption that no instantiation of variables has been done in the resolution of $C$ and $D$, and thus we have $A = \overline{B}$. We should also let $C - \{A\} = D - \{B\}$, which corresponds to an assumption that each literal in $R$ is inherited both from $C$ and $D$. Then we have

$$C = R \cup \{A\} \text{ and } D = R \cup \{\overline{A}\},$$

where $A$ could be any literal, and we say that $C$ and $D$ are obtained from $R$ by or-introduction of the literal $A$.

**Example** Consider the following clauses:
$$C = (\ p(f^2(a)) \leftarrow p(f(a)),\ p(a)\ ),$$
$$D = (\ p(f^2(a)), p(f(a)) \leftarrow p(a)\ ),\text{ and}$$
$$R = (\ p(f^2(a)) \leftarrow p(a)\ ).$$

The clauses $C$ and $D$ are parent clauses of $R$.

The above technique to find parent clauses can be used to reduce implication to $\theta$-subsumption. This is of interest for ambivalent clauses, such as $R$ in the example above, for which there are proper indirect roots. For example the clause
$$G = (\ p(f(x)) \leftarrow p(x)\ )$$
is a proper indirect second root of $R$, and it $\theta$-subsumes both $C$ and $D$.

Proposition 18 shows that the set of two clauses obtained by our technique for inverting one resolution is logically equivalent to the original clause. In Proposition 19 it is shown that or-introduction of one literal is a general technique for inverting one resolution.

**Proposition 18** *Let $R$ be a clause and $L$ a literal. Then $\{R\} \equiv \{R \cup \{L\}, R \cup \{\overline{L}\}\}$.*

**Proof:** Since $R \subseteq R \cup \{L\}$ and $R \subseteq R \cup \{\overline{L}\}$, we have $R \preceq R \cup \{L\}$ and $R \preceq R \cup \{\overline{L}\}$. Then by Proposition 3, $R \Rightarrow R \cup \{L\}$ and $R \Rightarrow R \cup \{\overline{L}\}$. Thus $\{R\} \models \{R \cup \{L\}, R \cup \{\overline{L}\}\}$. The clause $R$ is a resolvent of $R \cup \{L\}$ and $R \cup \{\overline{L}\}$. Then by soundness of resolution, $\{R \cup \{L\}, R \cup \{\overline{L}\}\} \models \{R\}$. Consequently, $\{R\} \equiv \{R \cup \{L\}, R \cup \{\overline{L}\}\}$. □

**Proposition 19** *Let $C$ and $D$ be clauses and $R$ a resolvent of $C$ and $D$. Then there exists a literal $L$ such that $C \preceq R \cup \{L\}$ and $D \preceq R \cup \{\overline{L}\}$.*

**Proof:** By the definition of a resolvent we have $R = ((C\gamma - \{A\}) \cup (D\mu - \{B\}))\theta$, where $C\gamma$ is a factor of $C$, $D\mu$ is a factor of $D$, $A \in C\gamma$, $B \in D\mu$ and $A\theta = \overline{B}\theta$ where $\theta$ is an mgu for $\{A, \overline{B}\}$. Let $L = A\theta$, and then $R \cup \{L\} = (C\gamma \cup (D\mu - \{B\}))\theta$, and $R \cup \{\overline{L}\} = ((C\gamma - \{A\}) \cup D\mu)\theta$ since $A\theta = \overline{B}\theta$. Hence $C\gamma\theta \subseteq (C\gamma \cup (D\mu - \{B\}))\theta$ and $D\mu\theta \subseteq ((C\gamma - \{A\}) \cup D\mu)\theta$, and consequently $C \preceq R \cup \{L\}$ and $D \preceq R \cup \{\overline{L}\}$. □

### 3.3 Inverting Multiple Resolutions by Or-introduction

The technique for inverting one resolution can be generalized to a technique for inverting a sequence of resolutions as follows. The set
$$\{R \cup \{L_1\}, R \cup \{\overline{L_1}\}\},$$
where $R$ is a clause and $L_1$ is a literal, is a set of two clauses from which the clause $R$ follows by one resolution. Similarly, the two sets
$$\{R \cup \{L_1\} \cup \{L_2\}, R \cup \{L_1\} \cup \{\overline{L_2}\}, R \cup \{\overline{L_1}\}\},\text{ and}$$
$$\{R \cup \{L_1\}, R \cup \{\overline{L_1}\} \cup \{L_2\}, R \cup \{\overline{L_1}\} \cup \{\overline{L_2}\}\},$$
where $R$ is a clause and $L_1$ and $L_2$ are literals, are sets of three clauses from which $R$ follows by two resolutions. In the same way, for given literals $L_1$, $L_2$ and $L_3$, we have six different sets of four clauses from which $R$ follows by three resolutions, and so on. All these sets are or-introduced from the clause $R$.

**Definition** Let $C$ be a clause and $\Omega$ a sequence of literals. Then a set of clauses $S$ is *or-introduced* from $C$ by $\Omega$ if and only if either:
a) $S = \{C\}$ and $\Omega = [\,]$, or
b) $S = (S' - \{D\}) \cup \{D \cup \{L\}, D \cup \{\overline{L}\}\}$ and $\Omega = [L_1, \ldots, L_n, L]$, where $S'$ is a set of clauses or-introduced from $C$ by $[L_1, \ldots, L_n]$ and $D \in S'$.

**Example** Consider the following clauses:
$$C = (\ p(f^3(a)) \leftarrow p(a)\ ),$$
$$D_1 = (\ p(f^3(a)), p(f^2(a)) \leftarrow p(a)\ ),$$
$$D_2 = (\ p(f^3(a)) \leftarrow p(f^2(a)), p(a)\ ),$$
$$E_1 = (\ p(f^3(a)), p(f^2(a)), p(f(a)) \leftarrow p(a)\ ),\ \text{and}$$
$$E_2 = (\ p(f^3(a)), p(f^2(a)) \leftarrow p(f(a)), p(a)\ ).$$

The set of clauses $\{D_1, D_2\}$ is or-introduced from $C$ by $[p(f^2(a))]$, and the set of clauses $\{E_1, E_2\}$ is or-introduced from $D_1$ by $[p(f(a))]$. Consequently, the set of clauses $\{D_2, E_1, E_2\}$ is or-introduced from $C$ by $[p(f^2(a)), p(f(a))]$.

In the example above, clause $D_1$ is a resolvent of $E_1$ and $E_2$, and $C$ is a resolvent of $D_1$ and $D_2$. Consequently, $C$ is derivable from $\{D_2, E_1, E_2\}$ by resolution. That a set of clauses or-introduced from a clause is logically equivalent to the clause, is shown by the following theorem.

**Theorem 20 (Equivalence preservation of or-introduction)** *Let $S$ be a set of clauses or-introduced from a clause $C$ by a sequence of literals $[L_1, \ldots, L_n]$. Then $S \equiv \{C\}$.*

**Proof:** The proof is by mathematical induction on $n$. It should be noted that $S$, in the statement of the theorem, in the proof is indexed by $n$.

*Base step (n=0):* $S_0$ is or-introduced from $C$ by $[\,]$. Hence $S_0 \equiv \{C\}$.

*Induction hypothesis (n=k):* $S_k \equiv \{C\}$, where $S_k$ is or-introduced from $C$ by $[L_1, \ldots, L_k]$.

*Induction step (n=k+1):* Let $D \in S_k$. Then $S_{k+1} = (S_k - \{D\}) \cup \{D \cup \{L_{k+1}\}, D \cup \{\overline{L_{k+1}}\}\}$ is or-introduced from $C$ by $[L_1, \ldots, L_k, L_{k+1}]$. By Proposition 18, we have $\{D \cup \{L_{k+1}\}, D \cup \{\overline{L_{k+1}}\}\} \equiv \{D\}$, and consequently $S_{k+1} \equiv S_k$. By the induction hypothesis $S_k \equiv \{C\}$, and thus $S_{k+1} \equiv \{C\}$. □

In section 3.2 we showed that it is possible to invert one resolution by or-introduction of one literal. Below we show that it is possible to invert a sequence of resolutions by or-introduction of a sequence of literals.

**Lemma 21** *Let $D$ and $E$ be clauses, $\{C_1, \ldots, C_n\}$ a set of clauses, and $\{D_1, \ldots, D_n\}$ a set of clauses or-introduced from $D$, such that $D \preceq E$ and, for every $1 \leq i \leq n$, $C_i \preceq D_i$. Then there exists a set of clauses $\{E_1, \ldots, E_n\}$ or-introduced from $E$, such that for every $1 \leq i \leq n$, $C_i \preceq E_i$.*

**Proof:** Let $D_i$ be an arbitrary clause in $\{D_1, \ldots, D_n\}$. Then we have $D_i = D \cup \Lambda_i$ for some set of literals $\Lambda_i$, since $\{D_1, \ldots, D_n\}$ is or-introduced from $D$. Since $C_i \preceq D_i$, there exists a substitution $\theta_i$ such that $C_i \theta_i \subseteq D \cup \Lambda_i$. Since $D \preceq E$, there exists a substitution $\sigma$ such that $D\sigma \subseteq E$. Thus we have $(D \cup \Lambda_i)\sigma \subseteq (E \cup \Lambda_i \sigma)$, and consequently $C_i \theta_i \sigma \subseteq E \cup \Lambda_i \sigma$. Let $E_i = E \cup \Lambda_i \sigma$ and we have $C_i \preceq E_i$. □

**Theorem 22 (Inverting resolution using or-introduction)** *Let $T$ be a set of clauses, $D$ a clause in $\mathcal{L}^n(T)$. Then there exists a set of clauses $S$ or-introduced from $D$ such that for each $E \in S$ there exists a clause $C \in T$ such that $C \preceq E$.*

**Proof:** The proof is by complete mathematical induction on $n$. It should be noted that $D$ and $S$, in the statement of the theorem, in the proof are indexed by $n$.

*Base step (n=1)*: By the definition of $n$th resolution layer $\mathcal{L}^1(T) = T$, and thus $D_1 \in T$. We have that $S_1 = \{D_1\}$ is or-introduced from $D_1$ by the empty sequence of literals $\Omega_1 = []$. Hence, for $D_1 \in S_1$ there exists a clause $D_1 \in T$ such that $D_1 \preceq D_1$.

*Induction hypothesis (n=k)*: For every $1 \leq i \leq k$, there exists a set of clauses $S_i$ or-introduced from $D_i$ by some sequence of literals $\Omega_i = [L_1, \ldots, L_{i-1}]$ such that for each $E \in S_i$ there exists a clause $C \in T$ such that $C \preceq E$.

*Induction step (n=k+1)*: By the definition of $n$th resolution layer, $D_{k+1}$ is a resolvent of some clauses $D_m \in \mathcal{L}^m(T)$ and $D_p \in \mathcal{L}^p(T)$ such that $m + p = k$, $1 \leq m \leq k$ and $1 \leq p \leq k$. Then by Proposition 19, there exists a literal $L$ such that $D_m \preceq D_{k+1} \cup \{L\}$ and $D_p \preceq D_{k+1} \cup \{\overline{L}\}$.

By the induction hypothesis, there exists a set of clauses $S_m$ or-introduced from $D_m$ by some sequence of literals $\Omega_m = [A_1, \ldots, A_{m-1}]$ such that for each $E \in S_m$ there exists a clause $C \in T$ such that $C \preceq E$. Then by Lemma 21, there exists a set of clauses $S'_m$ or-introduced from $D_{k+1} \cup \{L\}$ by some sequence of literals $\Omega'_m = [A'_1, \ldots, A'_{m-1}]$ such that for each $E' \in S'_m$ there exists a clause $C \in T$ such that $C \preceq E'$.

By the induction hypothesis, there also exists a set of clauses $S_p$ or-introduced from $D_p$ by some sequence of literals $\Omega_p = [B_1, \ldots, B_{p-1}]$ such that for each $E \in S_p$ there exists a clause $C \in T$ such that $C \preceq E$. Then by Lemma 21, there exists a set of clauses $S'_p$ or-introduced from $D_{k+1} \cup \{\overline{L}\}$ by some sequence of literals $\Omega'_p = [B'_1, \ldots, B'_p]$ such that for each $E' \in S'_p$ there exists a clause $C \in T$ such that $C \preceq E'$.

Then it follows from the definition of or-introduction that $S_{k+1} = S'_m \cup S'_p$ is a set of clauses or-introduced from $D_{k+1}$ by $\Omega_{k+1} = [L, A'_1, \ldots, A'_m, B'_1, \ldots, B'_p]$. Consequently, there exists a set of clauses $S_{k+1}$ or-introduced from $D_{k+1}$ such that for each $E \in S_{k+1}$ there exists a clause $C \in T$ such that $C \preceq E$. $\square$

### 3.4 Expansion of Clauses

In the section 3.3 it was described how a reduction of generalization can be achieved by replacing a clause by a set of clauses. Here we show how this set of clauses equivalently can be described by a single clause, which we call an expansion of the original clause. By definition, if a clause $C$ $\theta$-subsumes every clause in a set of clauses $S$, then $C$ will also $\theta$-subsume an LGG$\theta$ of $S$. This leads us to our definition of expansion of clauses. The idea of expansion of clauses was first presented by Idestam-Almquist (1993c).

**Definition** Let $D$ be a clause and $\Omega$ a sequence of literals. Then a clause $E$ is an *expansion* of $D$ by $\Omega$ if and only if $E$ is an LGG$\theta$ of a set of clauses or-introduced from $D$ by $\Omega$.

**Example** Consider the following clauses:

$$C = (\ p(f(x)) \leftarrow p(x)\ ),$$
$$D = (\ p(f^3(a)) \leftarrow p(a)\ ),$$
$$D_1 = (\ p(f^3(a)) \leftarrow p(f^2(a)),\ p(a)\ ),$$
$$D_2 = (\ p(f^3(a)),\ p(f^2(a)),\ p(f(a)) \leftarrow p(a)\ ),$$
$$D_3 = (\ p(f^3(a)),\ p(f^2(a)) \leftarrow p(f(a)),\ p(a)\ ),\ \text{and}$$
$$E = (\ p(f(x)),\ p(f^3(a)) \leftarrow p(a),\ p(x)\ ).$$

The set of clauses $\{D_1, D_2, D_3\}$ is or-introduced from the clause $D$ by $[p(f^2(a)), p(f(a))]$, and $E$ is an LGG$\theta$ of $\{D_1, D_2, D_3\}$. Consequently, $E$ is an expansion of $D$ by $[p(f^2(a)), p(f(a))]$.

Note that implication has been reduced to $\theta$-subsumption in the example above. We have $C \Rightarrow D$ and $C \not\preceq D$, but for the expansion $E$ of $D$ we have $C \preceq E$.

Expansion can be regarded as a transformation technique, since the expansion of a clause is logically equivalent to the clause itself.

**Theorem 23 (Equivalence preservation of expansion)** *Let $D$ be a clause, and $E$ an expansion of $D$. Then $E \Leftrightarrow D$.*

**Proof:** By the definition of expansion, we know that there exists a set of clauses $S$ or-introduced from $D$ by $\Omega$ such that $E$ is an LGG$\theta$ of $S$. By Theorem 20, we have $\{D\} \equiv S$. By the definition of an LGG$\theta$, $E \preceq F$ for each $F \in S$. Then by Proposition 3, $E \Rightarrow F$ for each $F \in S$. Thus $\{E\} \models S$, and consequently $E \Rightarrow D$.

We have $D \subseteq F$ for each $F \in S$. Then by the definition of an LGG$\theta$, we have $D \preceq E$, and by Proposition 3 $D \Rightarrow E$. Consequently, $E \Leftrightarrow D$. □

Below we prove that for every generalization under implication of a clause there exists an expansion of the clause such that the generalization under implication is reduced to a generalization under $\theta$-subsumption.

**Theorem 24 (Reduction of implication to $\theta$-subsumption using expansion)** *Let $C$ be a clause and $D$ a non-tautological clause such that $C \Rightarrow D$. Then there exists an expansion $E$ of $D$ such that $C \preceq E$.*

**Proof:** By Corollary 17, there exists a clause $D' \in \mathcal{L}^n(\{C\})$ such that $D' \preceq D$ for some $n \geq 1$. By Theorem 22, there exists a set of clauses $S'$ or-introduced from $D'$ such that for each $F' \in S'$ we have $C \preceq F'$. Then it follows from Lemma 21 that there exists a set of clauses $S$ or-introduced from $D$ such that for each $F \in S$ we have $C \preceq F$. Then let $E$ be an LGG$\theta$ of $S$, and thus an expansion of $D$, and we have $C \preceq E$ by the definition of an LGG$\theta$. □

### 3.5 Complete Expansion

Generalizations under implication of a clause can be reduced to generalizations under $\theta$-subsumption of an expansion of the clause. We are particularly interested in expansions of clauses such that every generalization under implication is reduced to a generalization under $\theta$-subsumption of that particular expansion.

**Definition** Let $D$ be a clause, and $E$ an expansion of $D$. Then $E$ is a *complete expansion* of $D$ if and only if, for every clause $C$, $C \preceq E$ whenever $C \Rightarrow D$.

Recall that an expansion of a clause is a clause, and thus finite. Muggleton and Page (1994, page 166) has shown that complete expansions, which they call finite self-saturations, do not exist for all clauses.

**Theorem 25 (Non-existence of complete expansions)** *There exist non-tautological clauses for which there exist no complete expansions.*

The non-existence of complete expansions is due to that for some clauses there are infinitely many distinct generalizations under implication. Because of this we turn to the problem of reducing every generalization under T-implication to a generalization under $\theta$-subsumption of a single expansion.

**Definition** Let $D$ be a clause, $E$ an expansion of $D$ and $T$ a term set of $\{D\}$. Then $E$ is a *T-complete expansion* of $D$ w.r.t. $T$ if and only if, for every clause $C$, $C \preceq E$ whenever $C \Rightarrow_T D$.

**Example** Consider the following clauses:

$$C_1 = (\ p(f(x)) \leftarrow p(x)\ ),$$
$$C_2 = (\ p(f^2(y)) \leftarrow p(y)\ ),$$
$$D = (\ p(f^4(a)) \leftarrow p(a)\ ),$$
$$E_1 = (\ p(f^2(y)),\ p(f^4(a)) \leftarrow p(a),\ p(y)\ ),\ \text{and}$$
$$E_2 = (\ p(f(x)),\ p(f^2(y)),\ p(f^4(a)) \leftarrow p(a),\ p(y),\ p(x)\ ).$$

The clauses $C_1$ and $C_2$ are proper indirect roots of $D$, such that $C_1 \Rightarrow_T D$ and $C_2 \Rightarrow_T D$. The clause $E_1$ is an expansion of $D$ by $[p(f^2(a))]$, and $E_1$ is an expansion of $D$ by $[p(f^2(a)),\ p(f^3(a)),\ p(f(a))]$. The expansion $E_2$ is a T-complete expansion but $E_1$ is not.

In the example above the T-complete expansion $E_2$ of $D$ is also a complete expansion of $D$. However, in contrast to complete expansions, T-complete expansions exist for all non-tautological clauses.

**Theorem 26 (Existence of T-complete expansions)** *Let $D$ be a non-tautological clause and $T$ a term set of $\{D\}$. Then there exists a T-complete expansion $E$ of $D$ w.r.t. $T$.*

**Proof:** By Theorem 13, there exists a complete LGGT $F$ of $\{D\}$ w.r.t. $T$. Hence, for every clause $C$, if $C \Rightarrow_T D$ then $C \preceq F$. By the definition of a complete LGGT, we have $F \Rightarrow_T D$, and then by Corollary 6, $F \Rightarrow D$. By theorem 24, there exists an expansion $E$ of $D$ such that $F \preceq E$. Thus, for every clause $C$, if $C \Rightarrow_T D$ then $C \preceq E$, and consequently $E$ is a T-complete expansion of $D$ w.r.t. $T$. $\square$

If we can compute T-complete expansions of a set of clauses then we can use Plotkin's algorithm for computing an LGG$\theta$ to compute an LGGT. From the proof of Theorem 26 it follows that the candidate set of literals to be used to compute a T-complete expansion is finite. Since expansion is equivalence preserving we could simply test all different ways to expand a clause by sequences of literals from this candidate set, and in this way obtain a T-complete expansion. This is of course an extremely complex process, but at least theoretically, T-complete expansions and LGGTs are computable.

## 4. Concluding Remarks

We have studied the problem of generalization of clauses. In section 2, we described the framework for generalization of clauses developed by Plotkin (1970, 1971b, 1971a), which is based on $\theta$-subsumption. Implication is the most natural basis for inductive generalization. In section 2, we therefore also studied the theory of generalization under implication.

The contents of section 2 can be summarized as follows:

1. It is decidable whether a clause $\theta$-subsumes another clause.

2. There exists a least general generalization under $\theta$-subsumption (LGG$\theta$) of every finite set of clauses.

3. It is undecidable whether a clause implies another clause.

4. It is an open problem whether there exists a least general generalization under implication (LGGI) of every finite set of clauses.

5. T-implication is a strictly stronger relation between clauses than implication, and strictly weaker than $\theta$-subsumption, and T-implication can become an arbitrarily good approximation of implication by extending the considered term set.

6. It is decidable whether a clause T-implies another clause.

7. There exists a least general generalization under T-implication (LGGT) of every finite set of clauses.

In section 3, we studied the difference between $\theta$-subsumption and implication on clauses. We presented our approach to find all generalizations under implication, by reducing implication to $\theta$-subsumption. This can be achieved by inverting self-resolution, and we described a technique for inverting resolution based on or-introduction of literals. We also described expansion of clauses, which summarizes our idea of reduction of implication to $\theta$-subsumption.

The contents of section 3 can be summarized as follows:

1. An expansion of a clause is an LGG$\theta$ of a set of clauses obtained by or-introduction from the clause.

2. For every generalization under implication of a clause there exists an expansion of the clause, logically equivalent to the clause, such that the generalization under implication is reduced to a generalization under $\theta$-subsumption.

3. There exist non-tautological clauses for which there exist no complete expansions, which means that there are no expansions of the clauses such that every generalization under implication is reduced to a generalization under $\theta$-subsumption of the expansions.

4. For each non-tautological clause there exists a T-complete expansion, which means that every generalization under T-implication of the clause is reduced to a generalization under $\theta$-subsumption of the expansion.

As noted in section 3.5, T-complete expansions and LGGTs are computable, but such a computation is extremely costly. This is not surprising since our framework for generalization under implication is based on and extends Plotkin's framework for generalization under $\theta$-subsumption, which already suffers from complexity problems. In general an LGG of a set of clauses may grow exponentially in the number of clauses in the set (Muggleton & Feng, 1990). Even an LGG reduced under $\theta$-subsumption, which means that all literals that are redundant under $\theta$-subsumption are removed, may grow exponentially in the number of clauses (Kietz, 1993). Since an expansion of a clause is an LGG of a set of or-introduced clauses, the computational cost of an expansion grows exponentially in the number of literals used in the or-introduction. In the computation of a T-complete expansion a large number of literals may be considered, and consequently such a computation would be extremely costly.

However, although Plotkin's framework for generalization under $\theta$-subsumption is computationally expensive, it has been widely used as a theoretical framework. Then to make it practical, a number of different restrictions on the clausal language has been considered, for example $ij$-determinacy (Muggleton & Feng, 1990). In a similar way we hope to find restrictions under which our here presented framework for generalization under implication can be practically useful.

Idestam-Almquist (1993b, 1993a) has described a technique to efficiently compute a restricted form of generalizations under implication. Recently, Muggleton has presented another approach based on generating a number of clauses, so called sub-saturants, which are candidates for being indirect roots, and then testing whether they are so or not (Muggleton, 1995). This approach might be a way to more efficiently compute some generalizations under implication. Some approaches to learn recursive definitions (recursive logic programs) by generalization under implication have been presented (Lapointe & Matwin, 1992; Aha, Lapointe, Ling, & Matwin, 1994; Idestam-Almquist, 1995). These approaches are based on structural analysis of the given examples, but can theoretically be described in our framework.

A study by Cohen (1995a, 1995b) of the learnability of recursive logic programs has previously been presented in this journal. In this study it was shown that a recursive logic program consisting of one constant-depth determinate closed $k$-ary recursive clause and one constant-depth determinate non-recursive clause is PAC-learnable given an additional "base-case oracle", which determines if a positive example is covered by the non-recursive base clause of the target program alone. It was also shown that generalizing this class of learning problem in any natural way leads to a computationally difficult problem. This result tells us that to efficiently learn more complex recursive hypotheses some extra information, such as rule models (Kietz & Wrobel, 1992) or program recursion schemes (Hamfelt & Nilsson, 1994), is needed.

The contributions of this paper are threefold. First, we have systematically reviewed and discussed the concepts relevant to generalization in a first-order setting. Second, we have introduced T-implication, a stronger form of implication which is decidable between clauses. Third, we have further developed previous work of the author (Idestam-Almquist, 1993c) on extending Plotkin's framework for generalization under $\theta$-subsumption to generalization under implication.

## Acknowledgements

The author wish to thank Torkel Franzén for invaluable help concerning the work on T-implication. The author also wish to thank Shan-Hwei Nienhuys-Cheng, Ronald de Wolf and the anonymous reviewers for a number of thoughtful comments and suggestions of improvements.

This work has been supported by the Swedish Research Council for Engineering Sciences (TFR) and the European Community ESPRIT BRA 6020 Inductive Logic Programming.